\documentclass[11pt,a4paper]{article}
\usepackage[hyperref]{emnlp2020}
\usepackage{times}
\usepackage{latexsym}
\usepackage{amsmath}
\usepackage{amsfonts,amssymb}

\usepackage{booktabs}
\usepackage{adjustbox}
\usepackage{graphicx}
\usepackage{subfigure}
\usepackage{multirow}
\usepackage{microtype}
\usepackage{pifont}
\definecolor{brickred}{rgb}{0.8, 0.25, 0.33}
\definecolor{brickred2}{rgb}{0.25, 0.8, 0.33}
\newcommand{\xm}{\color{brickred2}{\ding{51}}}%
\aclfinalcopy 
\def\aclpaperid{2083} 

\usepackage{tabularx,colortbl,xcolor}

\newcommand\hb{ \rowcolor{orange!15}}
\newcommand\hc{ \rowcolor{orange!40}}

\title{MAF: Multimodal Alignment Framework for \\ Weakly-Supervised Phrase Grounding}

\author{
Qinxin Wang$^{1}$, Hao Tan$^{2}$, Sheng Shen$^{3}$, Michael W. Mahoney$^{3}$, Zhewei Yao$^{3}$ \\
$^{1}$Shanghai Jiao Tong University~~~~~$^{2}$UNC Chapel Hill~~~~~$^{3}$University of California, Berkeley \\
\small\texttt{qinzzz@sjtu.edu.cn, haotan@cs.unc.edu, \{sheng.s, mahoneymw, zheweiy\}@berkeley.edu }
}

\date{\today}

\begin{document}
\maketitle


\begin{abstract}

Phrase localization is a task that studies the mapping from textual phrases to regions of an image.
Given difficulties in annotating phrase-to-object datasets at scale, we develop a Multimodal Alignment Framework (MAF) to leverage more widely-available caption-image datasets, which can then be used as a form of weak supervision.
We first present algorithms to model phrase-object relevance by leveraging fine-grained visual representations and visually-aware language representations.
By adopting a contrastive objective, our method uses information in caption-image pairs to boost the performance in weakly-supervised scenarios.
Experiments conducted on the widely-adopted Flickr30k dataset show a significant improvement over existing weakly-supervised methods.
With the help of the visually-aware language representations, we can also improve the previous best unsupervised result by 5.56\%.
We conduct ablation studies to show that both our novel model and our weakly-supervised strategies significantly contribute to our strong results.%
\footnote{Code is available at \url{https://github.com/qinzzz/Multimodal-Alignment-Framework}.}

\end{abstract}

\section{Introduction}
\label{sec:intro}

Language grounding involves mapping language to real objects or data.
Among language grounding tasks, phrase localization---which maps phrases to regions of an image---is a fundamental building block for other tasks. 
In the \emph{phrase localization task}, each data point consists of one image and its corresponding caption, i.e., $d = \left\langle I,S\right \rangle$, where $I$ denotes an image and $S$ denotes a caption. 
Typically, the caption $S$ contains several query phrases $\mathcal{P}= \left\{p_n \right\}^{N}_{n=1}$, where each phrase is grounded to a particular object in the image. 
The goal is to find the correct relationship between (query) phrases in the caption and particular objects in the image. 
Existing work~\cite{DBLP:journals/corr/RohrbachRHDS15, Kim2018BilinearAN, li2019visualbert, yu2018rethining, liu2019learning} mainly focuses on the supervised phrase localization setting.
This requires a large-scale annotated dataset of phrase-object pairs for model training. 
However, given difficulties associated with manual annotation of objects, the size of grounding datasets is often limited.
For example, the widely-adopted Flickr30k~\cite{10.1007/s11263-016-0965-7} dataset has 31k images, while the caption dataset MS COCO~\cite{lin2014microsoft} contains 330k images.

\begin{figure}[t]
    \centering
    \includegraphics[width=\linewidth]{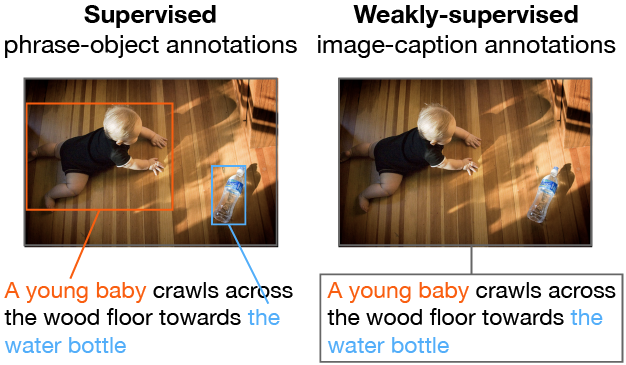}
    \caption{Comparison of phrase localization task under supervision (left) and weak supervision (right). 
    }
    \label{fig:sup_task}
    \vspace{-10pt}
\end{figure}

\begin{figure}[t]
\centering
    \includegraphics[width=.96\linewidth]{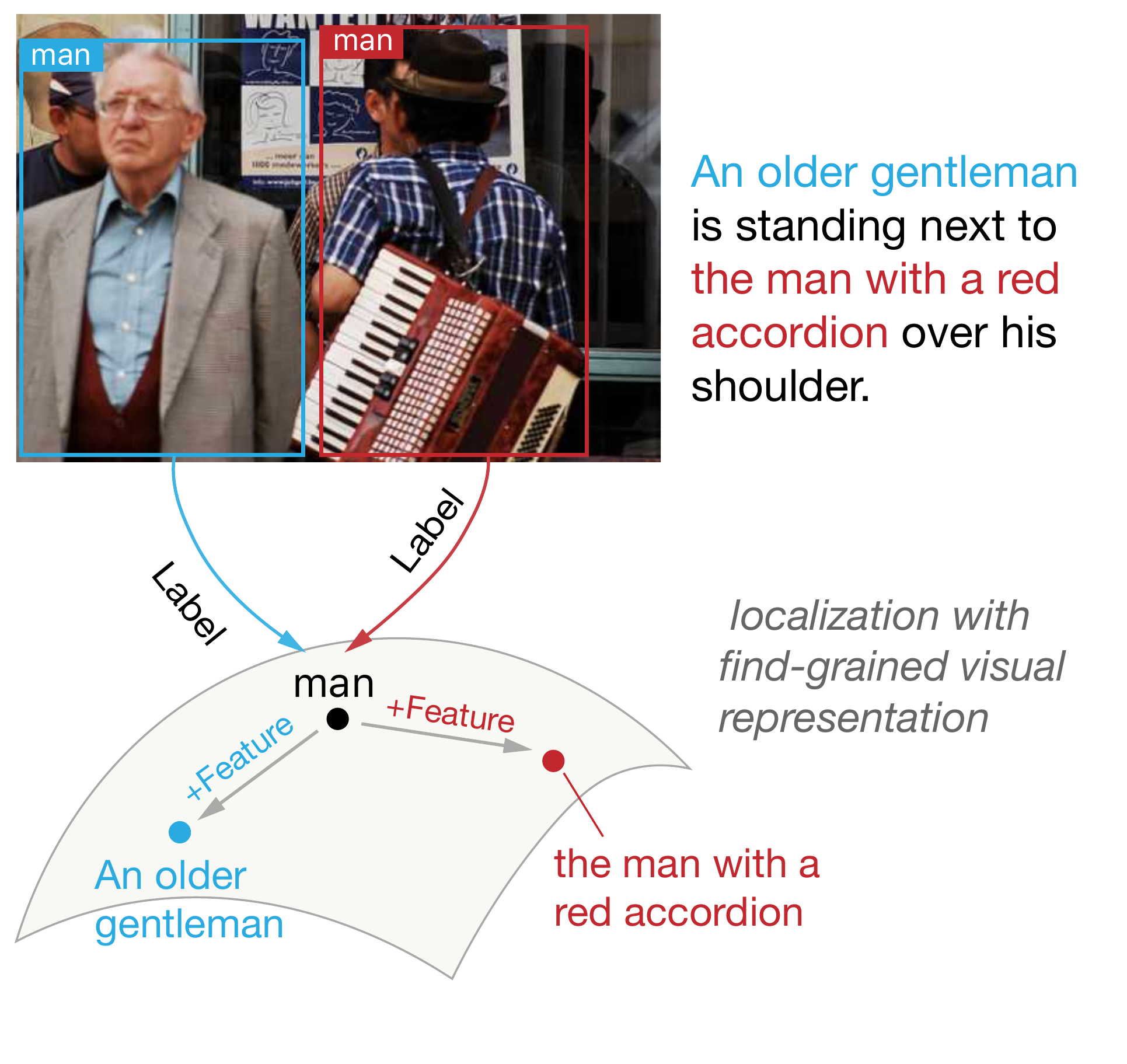}
        \caption{
        Example of the ambiguity caused by label-based localization (top); and our fine-grained visual representation disambiguate labels (bottom).}
        \label{feature}
        \vspace{-10pt}
\end{figure}

To address this limited data challenge, two different approaches have been proposed.
First, a weakly-supervised setting---which requires only \emph{caption-image annotations}, i.e., no \emph{phrase-object annotations}---was proposed by~\citet{DBLP:journals/corr/RohrbachRHDS15}.
This is illustrated in Figure~\ref{fig:sup_task}.
Second, an unsupervised setting---which does not need any training data, i.e., neither caption-image and phrase-object annotation---was proposed by~\citet{wang2019phrase}. 
To bring more semantic information in such a setting, previous work~\cite{yeh2018unsupervised, wang2019phrase} used the detected object labels from an off-the-shelf object detector (which we will generically denote by PreDet) and achieved promising results.
In more detail, for a given image $I$, the PreDet first generates a set of objects $\mathcal{O} = \left\{ o_m \right\}^{M}_{m=1}$.
Afterward, all the query phrases $\mathcal{P}$ and the detected objects $\mathcal{O}$ are fed into an alignment model to predict the final phrase-object pairs.
However, purely relying on the object labels causes ambiguity. 
For example, in Figure~\ref{feature}, the grounded objects of phrases ``an older man'' and ``the man with a red accordion'' are both labeled as ``man,'' and thus they are hard to differentiate.

\begin{figure*}[!htb]
    \centering
    \includegraphics[width=.96\linewidth]{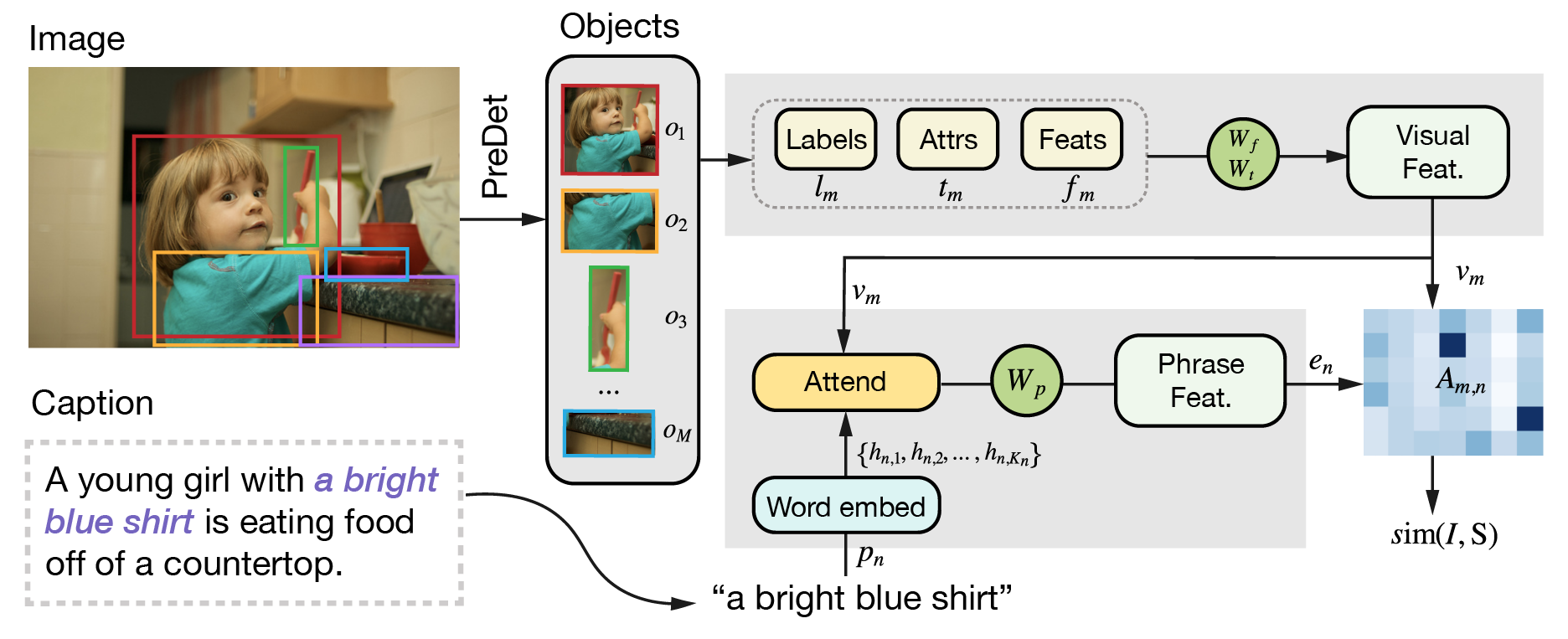}
    \caption{
    Overview of our proposed Multimodal Alignment Framework (MAF). A dataset of images and their captions  is the input to our model. PreDet predicts bounding boxes for objects in the image and their labels, attributes, and features, which are then integrated into visual feature representations. Attention is applied between word embedding and visual representations to compute the visually-aware language representations for phrases. Finally, a  multi-modal similarity function is used to measure the caption-image relevance based on the phrase-object similarity~matrix. }
    \label{model}
    \vspace{-5pt}
\end{figure*}

Given these observations, we propose a Multimodal Alignment Framework (MAF), which is illustrated in Figure~\ref{model}.
Instead of using only the label features from the PreDet (in our case, a Faster R-CNN~\cite{ren2015faster,anderson2018bottom}),
we also enhance the visual representations by integrating visual features from the Faster R-CNN into object labels.
(This is shown in Figure~\ref{feature}.) 
Next, we build visually-aware language representations for phrases, which thus could be better aligned with the visual representations.
Based on these representations, we develop a multimodal similarity function to measure the caption-image relevance with phrase-object matching scores.
Furthermore, we use a training objective to score relevant caption-image pairs higher than irrelevant caption-image pairs, which guides the alignment between visual and textual representations. 

We evaluate MAF on the public phrase localization dataset, Flickr30k Entities~\cite{10.1007/s11263-016-0965-7}.
Under the weakly-supervised setting (i.e., using only caption-image annotations without the more detailed phrase-object annotations), our method achieves an accuracy of 61.43\%, outperforming the previous weakly-supervised results by 22.72\%.
In addition, in the unsupervised setting, our visually-aware phrase representation improves the performance from the previous 50.49\% by 5.56\% up to 56.05\%.
Finally, we validate the effectiveness of model components, learning methods, and training techniques by showing their contributions to our final results.
\section{Related Work}
With the recent advancement in research in computer vision and computational linguistics, multimodal learning, which aims to explore the explicit relationship across vision and language, has drawn significant attention.  
Multimodal learning involves diverse tasks such as 
Captioning~\cite{vinyals2015show,xu2015show, karpathy2015deep, venugopalan2015sequence},
Visual Question Answering~\cite{anderson2018bottom, Kim2018BilinearAN, tan2019lxmert}, and 
Vision-and-Language Navigation~\cite{anderson2018vision, chen2019touchdown, thomason2020vision}.
Most of these tasks would benefit from better phrase-to-object localization, a task which attempts to learn a mapping between phrases in the caption and objects in the image by measuring their similarity. 
Existing works consider the phrase-to-object localization problem under various training scenarios, including supervised learning~\cite{DBLP:journals/corr/RohrbachRHDS15, yu2018rethining, liu2019learning, 10.1007/s11263-016-0965-7, li2019visualbert} and weakly-supervised learning~\cite{DBLP:journals/corr/RohrbachRHDS15, yeh2018unsupervised,  chen2018knowledge}.
Besides the standard phrase-object matching setup, previous works~\cite{DBLP:journals/corr/XiaoSL17, akbari2019multi, datta2019align2ground} have also explored a pixel-level ``pointing-game'' setting, which is easier to model and evaluate but less realistic.
Unsupervised learning was studied by~\citet{wang2019phrase}, who directly use word similarities between object labels and query phrases to tackle phrase localization without paired examples. 
Similar to the phrase-localization task, \citet{hessel2019unsupervised} leverages document-level supervision to discover image-sentence relationships over the~web.




\section{Methodology}


\subsection{Fine-grained Visual/Textual Features}

\paragraph{Visual Feature Representations.}
Previous works usually use only one specific output of the PreDet as the \emph{visual feature representation} (VFR).
For example, \citet{Kim2018BilinearAN} uses the final output feature of PreDet  (denoted as $\boldsymbol{f_m}$) as the VFR, 
and \citet{wang2019phrase} uses the label embedding (denoted as $\boldsymbol{l_m}$) of the predicted label from PreDet as the VFR. 
This unitary VFR usually lacks the counter-side information. 
Hence, we exploit different aspects of features extracted from PreDet for each object $o_m$ in the image. 
In particular, we consider the output feature $\boldsymbol{f_m}$, the label embedding $\boldsymbol{l_m}$, and the attribute embedding $\boldsymbol{t_m}$ of the object $o_m$ as the VFR,
\begin{equation}
    \boldsymbol{v_m} =  \boldsymbol{l_m} + W_t \boldsymbol{t_m} + W_f \boldsymbol{f_m}, \\ 
\end{equation}
where $W_t$ and $W_f$ are two projection matrices. 
Naively initializing $W_t$ and $W_f$ will lead the model to a sub-optimal solution. 
In Section~\ref{sec:empirical_results}, we discuss the effectiveness of different initializations.

\paragraph{Textual Feature Representations.}
\label{paragraph:tfr}
Existing works for \emph{textual feature representation} (TFR)~\cite{Kim2018BilinearAN, yu2018rethining, wang2019phrase} commonly treat it independently of the VFR. 
From a different angle, we use the attention between the textual feature and the VFR $\boldsymbol{v}_m$ to integrate the visual information from the object into TFR. 
In more detail, we first use the GloVe embedding~\cite{pennington2014glove} to encode the $K_n$ words in the phrase $p_n$ to $\left\{\boldsymbol{h}_{n,k}\right\}^{K_n}_{k=1}$, where$~\boldsymbol{h}_{n,k}\in \mathbb{R}^d$. 
Here, the dimension of $\boldsymbol{h}_{n,k}$ is the same as $\boldsymbol{v_m}$. 
We then define a word-object matching score $a_{n,k}^m$ for each $\boldsymbol{h_{n,k}}$ in the phrase to all object features $\boldsymbol{v_m}$. 
In particular, for each word $\boldsymbol{h}_{n,k}$ in the phrase, we select the object with the highest matching score, 
\begin{equation}
\begin{aligned}
    &\boldsymbol{a}_{n,k}^m = \mathrm{soft}\max_m \left\{\frac{\boldsymbol{h}_{n,k}^T\boldsymbol{v}_m}{{\sqrt{d}}}\right\},\\
    &\alpha_{n,k} = \max_{m} \{\boldsymbol{a}_{n,k}^m\}.
\end{aligned}
\end{equation}
Finally, we normalize the attention weights for each word in the phrase $p_n$ to obtain the final TFR, $\boldsymbol{e_n}$:
\begin{equation}
\begin{aligned}    
\beta_{n,k} &= \mathrm{soft}\max_k \left\{\alpha_{n,k}\right\},\\
\boldsymbol{e_n} &= W_p\left(\sum_k\beta_{n,k}\boldsymbol{h}_{n,k}\right).
\end{aligned}
\end{equation}
where $W_p$ is a projection matrix. 
In Section~\ref{sec:empirical_results}, we study the (superb) performance of the weight $\beta_{n,k}$ over simply the average $\boldsymbol{h}_{n,k}$ as well as the importance of the initialization of $W_p$.





\subsection{Training Objective and Learning Settings}
\label{sec:loss_setting}

\paragraph{Contrastive loss.}


For the weakly-supervised setting, 
we use a contrastive loss to train our model, due to the lack of phrase-object annotations. 
The contrastive objective $\mathcal{L}$ aims to learn the visual and textual features by maximizing the similarity score between paired image-caption elements and minimizing the score between the negative samples (i.e., other irrelevant images).
Inspired by the previous work in caption ranking~\cite{fang2015captions}, we use the following loss,
\begin{equation}
    \mathcal{L} = -\log \frac{e^{\text{sim}(I, S)}}{\sum_{I^{'} \in \mathit{batch}}{e^{\text{sim}(I^{'}, S)}}} .
\end{equation}
Here, $\text{sim}(I, S)$ is the similarity function defined below. 
Particularly, for each caption sentence, we use all the images $I^{'}$ in the current batch as candidate examples. 

\paragraph{Multimodal Similarity Functions.}
Following the document-level dense correspondence function in~\citet{hessel2019unsupervised}, our multimodal similarity function is defined as: 
\begin{equation}
\begin{aligned}
    \mathrm{sim}(I,S) = \frac{1}{N}\sum_{n}{\max_{m}{A_{n,m}}}.
\end{aligned}
\end{equation}
Here, $A \in \mathbb{R}^{N\times M}$ is the phrase-object similarity matrix, and its component is computed as
\begin{equation}
    A_{n,m} = \boldsymbol{e}_n^T \boldsymbol{v}_m, 
\end{equation}
and $\mathrm{sim}(I,S)$ measures the image-caption similarity.
It is calculated based on the similarity score between each phrase in the caption and each object in the image. 
Note that the maximum function $\max_{m}{A_{n,m}}$ directly connects our training objective and inference target, which alleviates the discrepancy between training and inference.

\paragraph{Weakly-supervised setting.} 
During training, our PreDet model is frozen. 
The word embeddings, $W_t$, $W_f$, and $W_p$ are trainable parameters. 
Here, the word embedding is initialized with GloVe~\cite{pennington2014glove}. 
We study the different initialization methods for the rest in Section~\ref{sec:empirical_results}. 
During inference, for the $n$-th phrase $p_n$ in an image-caption pair, we choose the localized object by
\begin{equation}
    m_{n}^\text{pred} = \arg\max_{m}{A_{n,m}} =  \arg\max_{m}{\boldsymbol{e}_n^T\boldsymbol{v}_{m}}.
\end{equation}

\paragraph{Unsupervised setting.}
\label{sec:unsup_setting}
In the unsupervised setting, the localized object is determined by
\begin{equation}
   m_{n}^\text{pred}  
   = \arg\max_m {\left(\sum_k\beta_{n,k}\boldsymbol{h}_{n,k}^{T}\right)\boldsymbol{l}_m}.
\end{equation}
We drop the parameters $W_t$, $W_f$, and $W_p$ here, because there is no training in the unsupervised setting.
$\beta_{n,k}$ is only calculated based on $\boldsymbol{l}_m$ (instead of $\boldsymbol{v}_m$).

\section{Empirical Results
}
\label{sec:empirical_results}


\paragraph{Dataset details.}
\label{sec:dataset_details}
The Flickr30k Entities dataset contains 224k phrases and 31k images in total, where each image will be associated with 5 captions and multiple localized bounding boxes. 
We use 30k images from the training set for training and 1k images for validation. 
The test set consists of 1k images with 14,481 phrases.
Our evaluation metric is the same as~\citet{10.1007/s11263-016-0965-7}.\footnote{To be specific, we use the evaluation code provided by~\newcite{wang2019phrase} at \url{https://github.com/josiahwang/phraseloceval}.} 
We consider a prediction to be correct if the IoU (Intersection of Union) score between our predicted bounding box and the ground-truth box is larger than 0.5. 
Following~\citet{DBLP:journals/corr/RohrbachRHDS15}, if there are multiple ground-truth boxes, we use their union regions as a single ground-truth bounding box for~evaluation.

\paragraph{Weakly-supervised Results.}
We report our weakly-supervised results on the test split in Table~\ref{tab:result_weaklysuper}. 
We include here upper bounds (UB), which are determined by the correct objects detected by the object detectors (if available).
Our MAF with ResNet-101-based Faster R-CNN detector pretrained on Visual Genome (VG)~\cite{krishnavisualgenome} can achieve an accuracy of 61.43\%.
This outperforms previous weakly-supervised methods by 22.71\%, and it narrows the gap between weakly-supervised and supervised methods to 15\%.
We also implement MAF with a VGG-based Faster R-CNN feature extractor pretrained on PASCAL VOC 2007~\cite{everingham2010pascal}, following the setting in KAC~\cite{chen2018knowledge}, and we use the same bounding box proposals as our ResNet-based detector. We achieve an accuracy of 44.39\%, which is 5.68\% higher than existing methods, showing a solid improvement under the same backbone~model.

{
\begin{table}[!htb]
    \caption{
    Weakly-supervised experiment results on Flick30k Entities.
    (We abbreviate backbone visual feature model as ``Vis. Feature,'' and upper bound as~``UB.'')
    }
    \label{tab:result_weaklysuper}
    \begin{adjustbox}{width=.5\textwidth,center} 
    \centering
    \begin{tabular}{lccc}
    \toprule
    Method & Vis. Features& Acc. (\%) & UB\\
    \midrule
    \multicolumn{3}{l}{\textbf{Supervised}} \\
    GroundeR \cite{DBLP:journals/corr/RohrbachRHDS15} & VGG$_\text{det}$ & 47.81& 77.90\\ 
    CCA~\cite{10.1007/s11263-016-0965-7} & VGG$_\text{det}$ & 50.89 & 85.12\\
    BAN \cite{Kim2018BilinearAN} & ResNet-101 & 69.69 & 87.45\\ 
    visualBERT \cite{li2019visualbert} & ResNet-101 &71.33& 87.45\\ 
    DDPN~\cite{yu2018rethining} & ResNet-101&73.30& - \\ 
    CGN~\cite{liu2019learning} &ResNet-101&76.74& - \\
    \midrule
    \multicolumn{3}{l}{\textbf{Weakly-Supervised}}\\
    GroundeR \cite{DBLP:journals/corr/RohrbachRHDS15} & VGG$_\text{det}$&28.93&77.90\\ 
    Link~\cite{yeh2018unsupervised} & YOLO$_\text{det}$ & 36.93& - \\ 
    KAC \cite{chen2018knowledge} & VGG$_\text{det}$ &38.71& - \\ 
    \midrule 
    \hb MAF (Ours) & VGG$_\text{det}$ & 44.39 & 86.29 \\ 
    \hc MAF (Ours) & ResNet-101 & \textbf{61.43} & 86.29\\
    \bottomrule
    \end{tabular}
    \end{adjustbox}
    \vspace{-10pt}
\end{table}
}

\paragraph{Unsupervised Results.\footnote{More unsupervised results are available in Appendix~\ref{appendix:baseline}.}}

We report our unsupervised results for the phrase localization method (described in Section~\ref{sec:unsup_setting}) in Table~\ref{tab:result_unsupervised}. 
For a fair comparison, we re-implemented~\citet{wang2019phrase} with a Faster R-CNN model trained on Visual Genome~\cite{krishnavisualgenome}.
This achieves 49.72\% accuracy (similar to 50.49\% as reported in their paper). 
Overall, our result (with VG detector) significantly outperforms the previous best result by 5.56\%, which demonstrates the effectiveness of our visually-aware language representations. 

{
\begin{table}[!htb]
    \large
    \caption{
    Unsupervised experiment results on Flick30k Entities. 
    w2v-max refers to the similarity algorithm proposed in~\cite{wang2019phrase}; Glove-att refers to our unsupervised inference strategy in Section~\ref{sec:loss_setting};
    CC, OI, and PL stand for detectors trained on MS COCO~\cite{lin2014microsoft}, Open Image~\cite{krasin2017openimages}, and Places~\cite{zhou2017places}.
    }
    \label{tab:result_unsupervised}
    \begin{adjustbox}{width=.5\textwidth,center} 
    \begin{tabular}{lccc}
    \toprule
    Method& TFR &Detector & Acc. (UB) (\%)\\
    \midrule
     Whole Image & None & None & 21.99 \\
    \cite{wang2019phrase} & w2v-max & Faster R-CNN & 49.72 (86.29)\\ 
    \cite{wang2019phrase} & w2v-max & CC+OI+PL & 50.49 (57.81)\\
    \midrule
    \hc MAF (Ours) & Glove-att & Faster R-CNN & 56.05 (86.29)\\ 
    \bottomrule
    \end{tabular}
        \end{adjustbox}
    \vspace{-10pt}
\end{table}
}

\paragraph{Ablation Experiments.}

In this section, we study the effectiveness of each component and learning strategy in MAF.
The comparison of different feature representations is shown in Table~\ref{tab:ablation_representation}. 
Replacing the visual attention based TFR with an average pooling based one
decreases the result from 61.43\% to lower than 60\%. 
For the VFR, using only object label $\boldsymbol{l}_m$ or visual feature $\boldsymbol{f}_m$ decreases the accuracy by 4.20\% and 2.94\%%
, respectively.
One interesting finding here is that the performance with all visual features (last row) is worse than the model with only $\boldsymbol{l}_m$ and $\boldsymbol{f}_m$.
Actually, we can infer that attributes cannot provide much information in localization (24.08\% accuracy if used alone), partly because attributes are not frequently used to differentiate objects in Flickr30k captions.

\begin{table}[!htb]
    \small
    \caption{ Ablation experiment results 
    of different visual and textual features. 
    TFR and VFR denotes textual and visual feature representation
    respectively.
    }
    \label{tab:ablation_representation}
    \centering
    \begin{adjustbox}{width=0.44\textwidth} 
    \begin{tabular}[t]{lcccccccccccccccccc}
    \toprule
    \multirow{2}{*}{TFR} & \multicolumn{3}{c}{VFR} & \multirow{2}{*}{Accuracy(\%)}\\
    \cmidrule{2-4}
        & $l_m$ & $f_m$ & $t_m$ & \\
    \midrule
    Average   & \xm & &  & 55.73 \\
    Average & & \xm & & 56.18\\
    Average & \xm &\xm &  & 59.51 \\
    \midrule
    Attention & \xm & &  &57.23 \\
    Attention & &\xm &  & 58.49 \\
    Attention & & &\xm & 24.08 \\ 
    Attention & \xm & &\xm  & 53.20 \\
    Attention & & \xm & \xm & 57.98 \\
    \hc Attention & \xm &\xm &  & 61.43 \\
    Attention & \xm &\xm &\xm  & 60.86 \\
    \bottomrule
    \end{tabular}
    \end{adjustbox}
    \vspace{-5pt}
    
\end{table}

\begin{table}[!htb]
    \caption{
    Ablation results of different initialization. 
    (ZR: zero initialization; RD: random initialization; ID+RD: noisy identity initialization.)
    }
    \small
    \label{tab:ablation_init} 
    \centering
    \begin{adjustbox}{width=0.44\textwidth} 
    \begin{tabular}[t]{cc|ccccccccccccccccccccccccccc}
    \toprule
    \multicolumn{2}{c}{$W_f$} & \multicolumn{2}{c}{$W_p$} &\multirow{2}{*}{Accuracy $\pm$ Var.(\%)}\\
    \cmidrule{1-4}
        ZR & RD & ID+RD & RD   & \\
    \midrule
      & \xm &  & \xm & 58.54 $\pm$ 0.26\\
    \midrule
    \xm &  && \xm& 60.05 $\pm$ 0.31\\
     &\xm  &\xm& & 59.68 $\pm$ 0.35\\
     \midrule
     \hc \xm&  &\xm& & 61.28 $\pm$ 0.32\\
    \bottomrule
    \end{tabular}
    \end{adjustbox}
    \vspace{-5pt}
\end{table}

We then investigate the effects of different initialization methods for the two weight matrices, $W_f$ and $W_p$. 
The results are presented in Table~\ref{tab:ablation_init}. 
Here ZR means zero initialization, RD means random initialization with Xavier~\cite{glorot2010understanding}, and ID+RD means identity with small random noise initialization.
We run each experiment for five times with different random seeds and compute the variance.
According to Table~\ref{tab:ablation_init}, the best combination is zero initialization for $W_f$ and identity+random initialization for $W_p$.
The intuitions here are: (i) For $W_f$, the original label feature $\boldsymbol{l}_m$ can have a non-trivial accuracy 57.23\% (see Table~\ref{tab:ablation_representation}), thus using RD on initializing $W_f$ will disturb the feature from $\boldsymbol{l}_m$;
(ii) For $W_p$, an RD initialization will disrupt the information from the attention mechanism, while ID+RD can both ensure basic text/visual feature matching and introduce a small random noise for training.
\section{Conclusions}
\label{sec:conclusions}
We present a Multimodal Alignment Framework, a novel method with fine-grained visual and textual representations for phrase localization, and we train it under a weakly-supervised setting, using a contrastive objective to guide the alignment between visual and textual representations.
We evaluate our model on Flickr30k Entities and achieve substantial improvements over the previous state-of-the-art methods with both weakly-supervised and unsupervised training strategies.
Detailed analysis is also provided to help future works investigate other critical feature enrichment and alignment methods for this task.



\section*{Acknowledgments}
We thank reviewers and the area chair for their helpful suggestions.
We are grateful for a gracious fund from the Amazon AWS.
MWM would also like to acknowledge DARPA, NSF, and ONR for providing partial support of this work.
HT is supported by Bloomberg Data Science Ph.D. Fellowship.
The views, opinions, and/or findings contained in this article are those of the authors and should not be interpreted as representing the official views or policies, either expressed or implied, of the funding agency.
\bibliographystyle{acl_natbib}
\bibliography{ref}

\begin{thebibliography}{32}
\expandafter\ifx\csname natexlab\endcsname\relax\def\natexlab#1{#1}\fi

\bibitem[{Akbari et~al.(2019)Akbari, Karaman, Bhargava, Chen, Vondrick, and
  Chang}]{akbari2019multi}
Hassan Akbari, Svebor Karaman, Surabhi Bhargava, Brian Chen, Carl Vondrick, and
  Shih-Fu Chang. 2019.
\newblock Multi-level multimodal common semantic space for image-phrase
  grounding.
\newblock In \emph{Proceedings of the IEEE Conference on Computer Vision and
  Pattern Recognition}, pages 12476--12486.

\bibitem[{Anderson et~al.(2018{\natexlab{a}})Anderson, He, Buehler, Teney,
  Johnson, Gould, and Zhang}]{anderson2018bottom}
Peter Anderson, Xiaodong He, Chris Buehler, Damien Teney, Mark Johnson, Stephen
  Gould, and Lei Zhang. 2018{\natexlab{a}}.
\newblock Bottom-up and top-down attention for image captioning and visual
  question answering.
\newblock In \emph{Proceedings of the IEEE conference on computer vision and
  pattern recognition}, pages 6077--6086.

\bibitem[{Anderson et~al.(2018{\natexlab{b}})Anderson, Wu, Teney, Bruce,
  Johnson, S{\"u}nderhauf, Reid, Gould, and van~den
  Hengel}]{anderson2018vision}
Peter Anderson, Qi~Wu, Damien Teney, Jake Bruce, Mark Johnson, Niko
  S{\"u}nderhauf, Ian Reid, Stephen Gould, and Anton van~den Hengel.
  2018{\natexlab{b}}.
\newblock Vision-and-language navigation: Interpreting visually-grounded
  navigation instructions in real environments.
\newblock In \emph{Proceedings of the IEEE Conference on Computer Vision and
  Pattern Recognition}, pages 3674--3683.

\bibitem[{Chen et~al.(2019)Chen, Suhr, Misra, Snavely, and
  Artzi}]{chen2019touchdown}
Howard Chen, Alane Suhr, Dipendra Misra, Noah Snavely, and Yoav Artzi. 2019.
\newblock Touchdown: Natural language navigation and spatial reasoning in
  visual street environments.
\newblock In \emph{Proceedings of the IEEE Conference on Computer Vision and
  Pattern Recognition}, pages 12538--12547.

\bibitem[{Chen et~al.(2018)Chen, Gao, and Nevatia}]{chen2018knowledge}
Kan Chen, Jiyang Gao, and Ram Nevatia. 2018.
\newblock Knowledge aided consistency for weakly supervised phrase grounding.
\newblock In \emph{Proceedings of the IEEE Conference on Computer Vision and
  Pattern Recognition}, pages 4042--4050.

\bibitem[{Datta et~al.(2019)Datta, Sikka, Roy, Ahuja, Parikh, and
  Divakaran}]{datta2019align2ground}
Samyak Datta, Karan Sikka, Anirban Roy, Karuna Ahuja, Devi Parikh, and Ajay
  Divakaran. 2019.
\newblock {Align2Ground}: Weakly supervised phrase grounding guided by
  image-caption alignment.
\newblock In \emph{Proceedings of the IEEE International Conference on Computer
  Vision}, pages 2601--2610.

\bibitem[{Everingham et~al.(2010)Everingham, Van~Gool, Williams, Winn, and
  Zisserman}]{everingham2010pascal}
Mark Everingham, Luc Van~Gool, Christopher~KI Williams, John Winn, and Andrew
  Zisserman. 2010.
\newblock The pascal visual object classes ({VOC}) challenge.
\newblock \emph{International journal of computer vision}, 88(2):303--338.

\bibitem[{Fang et~al.(2015)Fang, Gupta, Iandola, Srivastava, Deng, Doll{\'a}r,
  Gao, He, Mitchell, Platt et~al.}]{fang2015captions}
Hao Fang, Saurabh Gupta, Forrest Iandola, Rupesh~K Srivastava, Li~Deng, Piotr
  Doll{\'a}r, Jianfeng Gao, Xiaodong He, Margaret Mitchell, John~C Platt,
  et~al. 2015.
\newblock From captions to visual concepts and back.
\newblock In \emph{Proceedings of the IEEE conference on computer vision and
  pattern recognition}, pages 1473--1482.

\bibitem[{Glorot and Bengio(2010)}]{glorot2010understanding}
Xavier Glorot and Yoshua Bengio. 2010.
\newblock Understanding the difficulty of training deep feedforward neural
  networks.
\newblock In \emph{Proceedings of the thirteenth international conference on
  artificial intelligence and statistics}, pages 249--256.

\bibitem[{He et~al.(2016)He, Zhang, Ren, and Sun}]{he2016deep}
Kaiming He, Xiangyu Zhang, Shaoqing Ren, and Jian Sun. 2016.
\newblock Deep residual learning for image recognition.
\newblock In \emph{Proceedings of the IEEE conference on computer vision and
  pattern recognition}, pages 770--778.

\bibitem[{Hessel et~al.(2019)Hessel, Lee, and Mimno}]{hessel2019unsupervised}
Jack Hessel, Lillian Lee, and David Mimno. 2019.
\newblock Unsupervised discovery of multimodal links in multi-image,
  multi-sentence documents.
\newblock In \emph{Proceedings of the 2019 Conference on Empirical Methods in
  Natural Language Processing and the 9th International Joint Conference on
  Natural Language Processing (EMNLP-IJCNLP)}, pages 2034--2045.

\bibitem[{Karpathy and Fei-Fei(2015)}]{karpathy2015deep}
Andrej Karpathy and Li~Fei-Fei. 2015.
\newblock Deep visual-semantic alignments for generating image descriptions.
\newblock In \emph{Proceedings of the IEEE conference on computer vision and
  pattern recognition}, pages 3128--3137.

\bibitem[{Kim et~al.(2018)Kim, Jun, and Zhang}]{Kim2018BilinearAN}
Jin-Hwa Kim, Jaehyun Jun, and Byoung-Tak Zhang. 2018.
\newblock Bilinear attention networks.
\newblock In \emph{Advances in Neural Information Processing Systems}, pages
  1564--1574.

\bibitem[{Krasin et~al.(2017)Krasin, Duerig, Alldrin, Ferrari, Abu-El-Haija,
  Kuznetsova, Rom, Uijlings, Popov, Veit, Belongie, Gomes, Gupta, Sun, Chechik,
  Cai, Feng, Narayanan, and Murphy}]{krasin2017openimages}
Ivan Krasin, Tom Duerig, Neil Alldrin, Vittorio Ferrari, Sami Abu-El-Haija,
  Alina Kuznetsova, Hassan Rom, Jasper Uijlings, Stefan Popov, Andreas Veit,
  Serge Belongie, Victor Gomes, Abhinav Gupta, Chen Sun, Gal Chechik, David
  Cai, Zheyun Feng, Dhyanesh Narayanan, and Kevin Murphy. 2017.
\newblock Openimages: A public dataset for large-scale multi-label and
  multi-class image classification.
\newblock \emph{Dataset available from https://github.com/openimages}.

\bibitem[{Krishna et~al.(2017)Krishna, Zhu, Groth, Johnson, Hata, Kravitz,
  Chen, Kalantidis, Li, Shamma et~al.}]{krishnavisualgenome}
Ranjay Krishna, Yuke Zhu, Oliver Groth, Justin Johnson, Kenji Hata, Joshua
  Kravitz, Stephanie Chen, Yannis Kalantidis, Li-Jia Li, David~A Shamma, et~al.
  2017.
\newblock {Visual Genome}: Connecting language and vision using crowdsourced
  dense image annotations.
\newblock \emph{International Journal of Computer Vision}, 123(1):32--73.

\bibitem[{Li et~al.(2019)Li, Yatskar, Yin, Hsieh, and Chang}]{li2019visualbert}
Liunian~Harold Li, Mark Yatskar, Da~Yin, Cho-Jui Hsieh, and Kai-Wei Chang.
  2019.
\newblock {VisualBERT}: A simple and performant baseline for vision and
  language.
\newblock \emph{arXiv preprint arXiv:1908.03557}.

\bibitem[{Lin et~al.(2014)Lin, Maire, Belongie, Hays, Perona, Ramanan,
  Doll{\'a}r, and Zitnick}]{lin2014microsoft}
Tsung-Yi Lin, Michael Maire, Serge Belongie, James Hays, Pietro Perona, Deva
  Ramanan, Piotr Doll{\'a}r, and C~Lawrence Zitnick. 2014.
\newblock Microsoft {COCO}: Common objects in context.
\newblock In \emph{European conference on computer vision}, pages 740--755.
  Springer.

\bibitem[{Liu et~al.(2020)Liu, Wan, Zhu, and He}]{liu2019learning}
Yongfei Liu, Bo~Wan, Xiaodan Zhu, and Xuming He. 2020.
\newblock Learning cross-modal context graph for visual grounding.
\newblock In \emph{Proceedings of the AAAI Conference on Artificial
  Intelligenc}.

\bibitem[{Pennington et~al.(2014)Pennington, Socher, and
  Manning}]{pennington2014glove}
Jeffrey Pennington, Richard Socher, and Christopher~D. Manning. 2014.
\newblock Glove: Global vectors for word representation.
\newblock In \emph{Empirical Methods in Natural Language Processing (EMNLP)},
  pages 1532--1543.

\bibitem[{Plummer et~al.(2015)Plummer, Wang, Cervantes, Caicedo, Hockenmaier,
  and Lazebnik}]{10.1007/s11263-016-0965-7}
Bryan~A Plummer, Liwei Wang, Chris~M Cervantes, Juan~C Caicedo, Julia
  Hockenmaier, and Svetlana Lazebnik. 2015.
\newblock {Flickr30k Entities}: Collecting region-to-phrase correspondences for
  richer image-to-sentence models.
\newblock In \emph{Proceedings of the IEEE international conference on computer
  vision}, pages 2641--2649.

\bibitem[{Ren et~al.(2015)Ren, He, Girshick, and Sun}]{ren2015faster}
Shaoqing Ren, Kaiming He, Ross Girshick, and Jian Sun. 2015.
\newblock Faster {R-CNN}: Towards real-time object detection with region
  proposal networks.
\newblock In \emph{Advances in neural information processing systems}, pages
  91--99.

\bibitem[{Rohrbach et~al.(2016)Rohrbach, Rohrbach, Hu, Darrell, and
  Schiele}]{DBLP:journals/corr/RohrbachRHDS15}
Anna Rohrbach, Marcus Rohrbach, Ronghang Hu, Trevor Darrell, and Bernt Schiele.
  2016.
\newblock Grounding of textual phrases in images by reconstruction.
\newblock In \emph{European Conference on Computer Vision}, pages 817--834.
  Springer.

\bibitem[{Tan and Bansal(2019)}]{tan2019lxmert}
Hao Tan and Mohit Bansal. 2019.
\newblock {LXMERT}: Learning cross-modality encoder representations from
  transformers.
\newblock In \emph{Proceedings of the 2019 Conference on Empirical Methods in
  Natural Language Processing and the 9th International Joint Conference on
  Natural Language Processing (EMNLP-IJCNLP)}, pages 5103--5114.

\bibitem[{Thomason et~al.(2020)Thomason, Murray, Cakmak, and
  Zettlemoyer}]{thomason2020vision}
Jesse Thomason, Michael Murray, Maya Cakmak, and Luke Zettlemoyer. 2020.
\newblock Vision-and-dialog navigation.
\newblock In \emph{Conference on Robot Learning}, pages 394--406.

\bibitem[{Venugopalan et~al.(2015)Venugopalan, Rohrbach, Donahue, Mooney,
  Darrell, and Saenko}]{venugopalan2015sequence}
Subhashini Venugopalan, Marcus Rohrbach, Jeffrey Donahue, Raymond Mooney,
  Trevor Darrell, and Kate Saenko. 2015.
\newblock Sequence to sequence-video to text.
\newblock In \emph{Proceedings of the IEEE international conference on computer
  vision}, pages 4534--4542.

\bibitem[{Vinyals et~al.(2015)Vinyals, Toshev, Bengio, and
  Erhan}]{vinyals2015show}
Oriol Vinyals, Alexander Toshev, Samy Bengio, and Dumitru Erhan. 2015.
\newblock Show and tell: A neural image caption generator.
\newblock In \emph{Proceedings of the IEEE conference on computer vision and
  pattern recognition}, pages 3156--3164.

\bibitem[{Wang and Specia(2019)}]{wang2019phrase}
Josiah Wang and Lucia Specia. 2019.
\newblock Phrase localization without paired training examples.
\newblock In \emph{Proceedings of the IEEE International Conference on Computer
  Vision}, pages 4663--4672.

\bibitem[{Xiao et~al.(2017)Xiao, Sigal, and
  Jae~Lee}]{DBLP:journals/corr/XiaoSL17}
Fanyi Xiao, Leonid Sigal, and Yong Jae~Lee. 2017.
\newblock Weakly-supervised visual grounding of phrases with linguistic
  structures.
\newblock In \emph{Proceedings of the IEEE Conference on Computer Vision and
  Pattern Recognition}, pages 5945--5954.

\bibitem[{Xu et~al.(2015)Xu, Ba, Kiros, Cho, Courville, Salakhudinov, Zemel,
  and Bengio}]{xu2015show}
Kelvin Xu, Jimmy Ba, Ryan Kiros, Kyunghyun Cho, Aaron Courville, Ruslan
  Salakhudinov, Rich Zemel, and Yoshua Bengio. 2015.
\newblock Show, attend and tell: Neural image caption generation with visual
  attention.
\newblock In \emph{International conference on machine learning}, pages
  2048--2057.

\bibitem[{Yeh et~al.(2018)Yeh, Do, and Schwing}]{yeh2018unsupervised}
Raymond~A Yeh, Minh~N Do, and Alexander~G Schwing. 2018.
\newblock Unsupervised textual grounding: Linking words to image concepts.
\newblock In \emph{Proceedings of the IEEE Conference on Computer Vision and
  Pattern Recognition}, pages 6125--6134.

\bibitem[{Yu et~al.(2018)Yu, Yu, Xiang, Zhao, Tian, and Tao}]{yu2018rethining}
Zhou Yu, Jun Yu, Chenchao Xiang, Zhou Zhao, Qi~Tian, and Dacheng Tao. 2018.
\newblock Rethinking diversified and discriminative proposal generation for
  visual grounding.
\newblock \emph{International Joint Conference on Artificial Intelligence
  (IJCAI)}.

\bibitem[{Zhou et~al.(2017)Zhou, Lapedriza, Khosla, Oliva, and
  Torralba}]{zhou2017places}
Bolei Zhou, Agata Lapedriza, Aditya Khosla, Aude Oliva, and Antonio Torralba.
  2017.
\newblock Places: A 10 million image database for scene recognition.
\newblock \emph{IEEE transactions on pattern analysis and machine
  intelligence}, 40(6):1452--1464.

\end{thebibliography}

\appendix
\section{Implementation Details}
\label{appendix:Impl}
For GloVE word embeddings, we use the one with the hidden dimension 300. 
Phrases are split into words by space. 
We replace all out-of-vocabulary words with the introduced $\left\langle \text{UNK} \right\rangle$ token.
For object proposals, we apply an off-the-shelf Faster R-CNN model~\cite{ren2015faster} as the object detector\footnote{\small https://github.com/jwyang/faster-rcnn.pytorch} for object pseudo-labels. 
The backbone of the detector is ResNet-101~\cite{he2016deep}, and it is pre-trained on Visual Genome with mAP=10.1. 
We keep all bounding boxes with a confidence score larger than 0.1.
For ResNet-based visual features, we use the 2048-dimensional feature from Bottom-up attention \cite{anderson2018bottom}, which is pre-trained with 1600 object labels and 400 attributes.

The extracted visual features are frozen during training, and we use a batch size of 64 during training. 
Our optimizer is Adam with learning rate $lr = 1e^{-5}$. 
Except for word embeddings, trainable parameters include  $W_t \in \mathbb{R}^{d_T \times d_T}$, $W_f \in \mathbb{R}^{d_V \times d_T}$, and $W_p \in \mathbb{R}^{d_T \times d_T}$, where $d_T=300$, $d_V=2048$ for ResNet-101 backbone and $d_V=4096$ for VGG backbone. During training, it takes around 350 seconds to train an epoch using a single Tesla K80.
We train our model for 25 epochs and report the results at the last epoch.


\section{Baselines}
\label{appendix:baseline}
In Table~\ref{tab:result_baselines}, we report the results of different unsupervised methods:
\begin{itemize}
    \item Random: Randomly localize to a detected object.
    \item Center-obj: Localize to the object which is closest to the center of image, where we use an $L_1$ distance $D = |x-x_{\text{center}}| + |y-y_{\text{center}}|$.
    \item Max-obj: Localize to the object with the maximal area.
    \item Whole Image: Always localize to the whole image.
    \item Direct Match: Localize with the direct match between object labels and words in the phrase, e.g., localize ``a red apple'' to the object with the label ``apple.'' 
    If multiple labels are matched, we choose the one with the largest bounding~box.
    \item Glove-max: Consider every word-label similarity independently and select the object label with the highest semantic similarity with any~word.
    \item Glove-avg: Represent a phrase using an average pooling over Glove word embeddings and select the object label with highest the semantic similarity with the phrase representation.
    \item Glove-att: Use our visual attention based phrase representation, as is described in the Methodology~\ref{paragraph:tfr}.
\end{itemize}

Note that in all label-based methods (Direct Match \cite{wang2019phrase}, and our unsupervised method), if multiple bounding boxes share the same label, we choose the largest one as the predicted box.

{
\begin{table}[!htb]
    \caption{
    Baseline results of unsupervised methods on Flick30k Entities. Abbreviations are explained above.
    }
    \small
    \label{tab:result_baselines}
    \begin{adjustbox}{width=.45\textwidth,center} 
    \begin{tabular}{lccc}
    \toprule
    Method & Detector & Acc. (\%)\\
    \midrule
    Random & Faster R-CNN & 7.19\\
    Center-obj &Faster R-CNN& 18.24\\
    Whole Image &None & 21.99 \\
    Max-obj &Faster R-CNN& 24.51\\
    Direct match &Faster R-CNN & 26.42\\
    \midrule
    Glove-max & Faster R-CNN & 26.28\\
    Glove-avg & Faster R-CNN& 54.51\\
    Glove-att & Faster R-CNN & 56.05 \\
    
    \bottomrule
    \end{tabular}
        \end{adjustbox}
\end{table}
}

\section{Qualitative Analysis}
\label{appendix:analysis}
\begin{figure}[!bth]
    \centering
    \includegraphics[scale = 0.32]{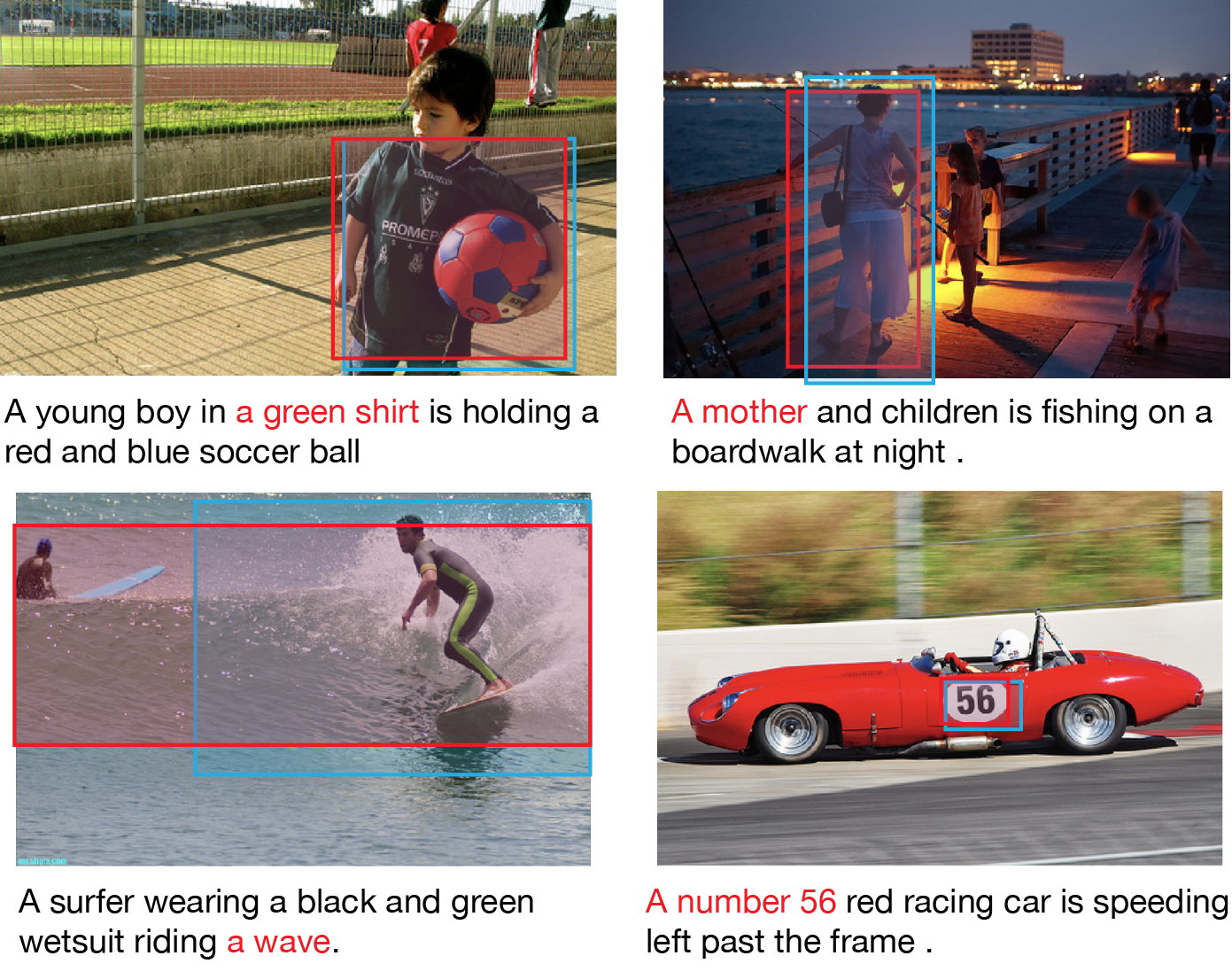}
    \caption{Example of predictions on Flickr30k. (Red box: ground truth, blue box: our prediction).}
    \label{fig:example1}
\end{figure}

\begin{figure}[!bth]
    \centering
    \hspace{5pt}
    \includegraphics[scale = 0.34]{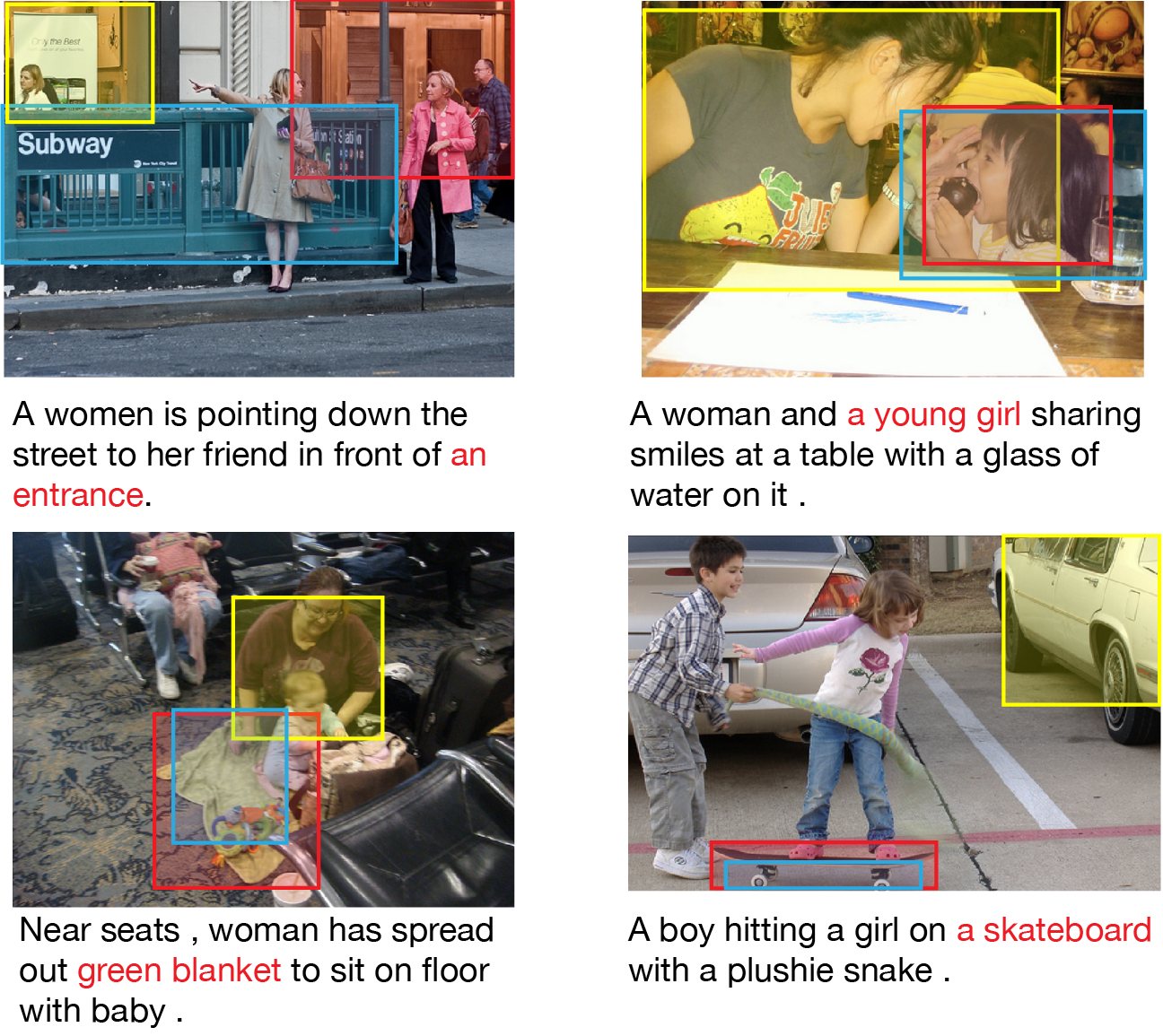}
    \caption{Example of predictions on Flickr30k. (Red box: ground truth, blue box: supervised prediction, yellow box: unsupervised prediction)}
    \label{fig:example2}
    \vspace{-5pt}
\end{figure}
To analyze our model qualitatively, we show some visualization results in Figure~\ref{fig:example1} and Figure~\ref{fig:example2}.
Figure~\ref{fig:example1} shows examples with consistent predictions between supervised and unsupervised models. 
In these cases, both methods can successfully learn to localize various objects, including persons (``mother''), clothes (``shirt''), landscapes (``wave''), and numbers (``56'').
Figure~\ref{fig:example2} shows examples where supervised and unsupervised methods localize to different objects. 
In the first image, they both localize the phrase ``entrance'' incorrectly. 
In the remaining three images, the supervised method learns to predict a tight bounding box on the correct object, while the unsupervised method localizes to other irrelevant objects.
For example (bottom left figure for Figure~\ref{fig:example2}), if the object detector fails to detect the ``blanket,'' then the unsupervised method can never localize ``green blanket'' to the right object. Still, the supervised method can learn from negative examples and obtain more information.

\end{document}